\documentclass[11pt]{article}

\usepackage{acl}
\usepackage{tcolorbox}
\usepackage{enumitem}
\usepackage{xcolor}

\definecolor{highlightred}{HTML}{D55E00}
\definecolor{choicegreen}{HTML}{009E73}
\usepackage{times}
\usepackage{latexsym}
\usepackage{hyperref}
\usepackage[T1]{fontenc}
\usepackage{float}
\usepackage[utf8]{inputenc}

\usepackage{microtype}

\usepackage{inconsolata}

\usepackage{graphicx}
\usepackage{booktabs}
\usepackage{multirow}
\usepackage{xcolor}
\usepackage{array}
\usepackage{rotating}
\usepackage{adjustbox}
\usepackage{graphicx}
\usepackage{subcaption}
\usepackage{caption}
\usepackage{amsmath}
\usepackage{amssymb}
\usepackage{tabularx} 
\title{Do Audio LLMs Really LISTEN, or Just Transcribe? \\ Measuring Lexical vs. Acoustic Emotion Cues Reliance}

\author{
\textbf{Jingyi Chen\textsuperscript{1,2}},
\textbf{Zhimeng Guo\textsuperscript{3}},
\textbf{Jiyun Chun\textsuperscript{2}},
\textbf{Pichao Wang\textsuperscript{4}},
\textbf{Andrew Perrault\textsuperscript{2}},
\textbf{Micha Elsner\textsuperscript{1}}
\\[4pt]
\textsuperscript{1}Department of Linguistics, The Ohio State University, USA\\
\textsuperscript{2}Department of Computer Science and Engineering, The Ohio State University, USA\\
\textsuperscript{3}Department of Information Sciences and Technology, Penn State University, USA\\
\textsuperscript{4}Amazon, USA\\[4pt]
\small{
\href{mailto:chen.9220@osu.edu}{chen.9220@osu.edu},
\href{mailto:chun.203@osu.edu}{chun.203@osu.edu},
\href{mailto:elsner.14@osu.edu}{elsner.14@osu.edu},
\href{mailto:perrault.17@osu.edu}{perrault.17@osu.edu},
\href{mailto:zzg5107@psu.edu}{zzg5107@psu.edu},
\href{mailto:pichaowang@gmail.com}{pichaowang@gmail.com}
}
}

\begin{document}
\maketitle
\footnotetext[4]{This work does not relate to the author's position at Amazon.}
\begin{abstract}
Understanding emotion from speech requires sensitivity to both lexical and acoustic cues. However, it remains unclear whether large audio language models (LALMs) genuinely process acoustic information or rely primarily on lexical contents. We present LISTEN (Lexical vs. Acoustic Speech Test for Emotion in Narratives), a controlled benchmark designed to disentangle lexical reliance from acoustic sensitivity in emotion understanding.
Across evaluations of six state-of-the-art LALMs, we observe a consistent lexical dominance. Models predict “neutral” when lexical cues are neutral or absent, show limited gains under cue alignment, and fail to classify distinct emotions under cue conflict. In paralinguistic settings, performance approaches chance.
These results indicate that current LALMs largely “transcribe” rather than “listen”, relying heavily on lexical semantics while underutilizing acoustic cues. LISTEN offers a principled framework for assessing emotion understanding in multimodal models. Project website: \url{https://delijingyic.github.io/LISTEN-website/}.
\end{abstract}

\section{Introduction}

Large audio language models (LALMs) have recently demonstrated impressive capabilities in multimodal reasoning, enabling systems to process spoken input and generate naturalistic responses. These advances hold particular promise for applications requiring social and emotional intelligence, where successful interaction depends not only on lexical content but also on acoustic cues such as pitch, intonation, and rhythm \cite{openai2024gpt4ocard, comanici2025gemini25pushingfrontier,xu2025qwen3omnitechnicalreport}. However, an open question remains: to what extent do these models actually make use of the speech signal itself, rather than relying on lexical cues alone?

This uncertainty arises in part from how LALMs are constructed. Most contemporary models are adapted from large text-only LLMs through multimodal fine-tuning with paired speech–text data. While this process transfers strong linguistic and reasoning abilities, it also raises the possibility of a structural bias: models may inherit a preference for lexical cues, while treating acoustic information as secondary. Such a bias is particularly problematic because speech is not merely text presented in an alternative modality. Beyond lexical content, spoken communication carries acoustic and paralinguistic signals, including intonation, pitch, intensity, rhythm, and voice quality, that are central to how meaning is conveyed \cite{scherer2003vocal,banse1996acoustic}. These cues frequently interact with, and in some cases override, the lexical channel. A salient example is sarcasm, where the intended emotional stance directly contradicts the literal words; listeners correctly interpret the speaker's intent by relying primarily on acoustic cues rather than lexical semantics \cite{bryant2005there}.

However, current benchmarks provide limited diagnostic insight into this issue. Many datasets contain emotionally explicit words (e.g., furious, delighted), which enable models to achieve high accuracy by exploiting transcript-based shortcuts. Consequently, strong performance on standard emotion recognition tasks may overestimate a model's ability to process acoustic and paralinguistic information, leaving open the question of whether these systems are genuinely \emph{listening} to speech.

To address this gap, we introduce \textbf{LISTEN}: Lexical vs. Acoustic Speech Test for Emotion in Narratives, a new benchmark explicitly designed to disentangle lexical reliance from acoustic sensitivity in emotion understanding. Our evaluation framework spans four controlled conditions that manipulate the relationship between lexical cues and acoustic cues:
(i) Neutral-Text, where lexical contents are emotionally neutral but acoustic cue varies, isolating the contribution of acoustic cues;
(ii) Emotion-Matched, where lexical and acoustic cues are aligned;
(iii) Emotion-Mismatched, where lexical and acoustic cues conflict, as in sarcasm; and
(iv) Paralinguistic, where affect is conveyed without lexical content (e.g., laughter, sighs).
Within Neutral-Text, Emotion-Matched, and Emotion-Mismatched conditions, we systematically compare performance across Text-only, Audio-only, and Text+Audio modalities. This design enables us to probe whether LALMs succeed by genuinely processing acoustic information or by defaulting to transcript-based shortcuts.

\paragraph{Our contributions} 
(1) We introduce LISTEN, the first diagnostic benchmark explicitly constructed to separate lexical and acoustic effects in emotion understanding through controlled cue manipulation and multimodal evaluation.
(2) We systematically evaluate six state-of-the-art open- and closed-weight LALMs across all conditions and modalities, revealing a consistent lexical dominance that limits true listening ability.
(3) We analyze how cue alignment, conflict, and absence each shape model behavior, offering new insight into why current audio language models “transcribe” emotion more than they “listen” to it.
LISTEN benchmark is available at: \url{https://huggingface.co/datasets/VibeCheck1/LISTEN_full}. Code is available at \url{https://github.com/DeliJingyiC/LISTEN}.

\section{Related work}
\begin{figure*}[h]
  \centering
    \centering
    \includegraphics[width=1\linewidth]{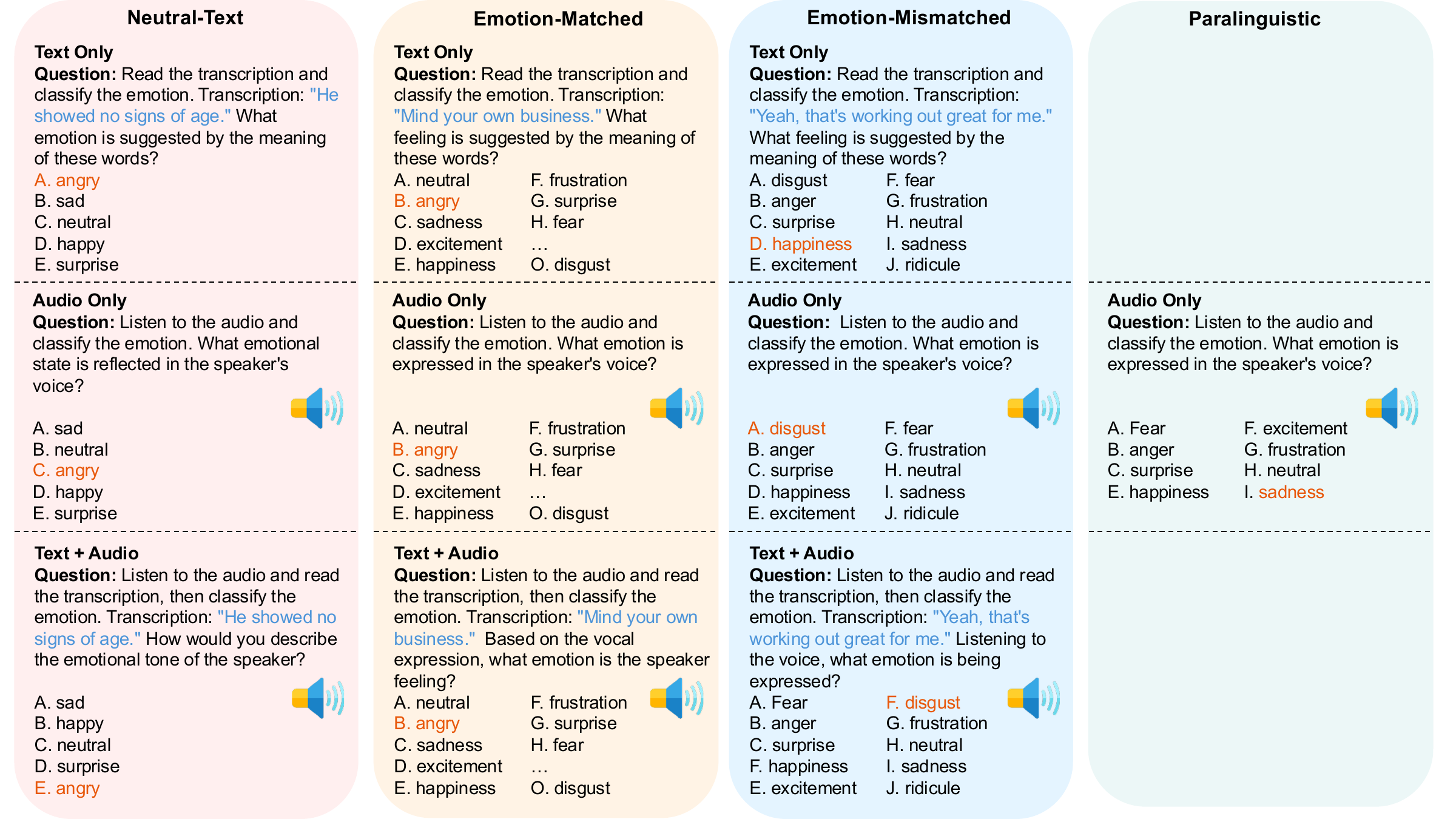}
  \caption{Examples from the LISTEN benchmark.}
  \label{fig:LISTEN_example}
\end{figure*}
\paragraph{Audio Benchmarks}
Recent benchmarks have broadened LALM evaluation across diverse audio domains. MMAU \cite{sakshi2024mmaumassivemultitaskaudio} introduced 10k QA pairs over 27 skills for speech, sound, and music, while MMAU-Pro \cite{kumar2025mmauprochallengingcomprehensivebenchmark} extended it to long-form, multi-source, and culturally diverse audio. MMAR \cite{ma2025mmarchallengingbenchmarkdeep} added 1k real-world QA triplets with hierarchical reasoning, though limited in scale. AudioBench \cite{wang-etal-2025-audiobench} unified 26 datasets in eight task types, and MuChoMusic \cite{weck2024muchomusicevaluatingmusicunderstanding} tested 1.1k MCQs highlighting textual bias. MMSU \cite{wang2025mmsu} evaluated 5k spoken QA pairs across 47 skills, and Dynamic-SUPERB Phase-2 \cite{huang2025dynamicsuperbphase2collaborativelyexpanding} covered 180 instruction-tuned speech, music, and sound tasks. AHELM \cite{lee2025ahelm} offered a holistic benchmark spanning perception, reasoning, emotion, bias, and multilingual safety.
Although several of these benchmarks include emotion or sarcasm tasks, none directly probe how models use the speech signal itself—whether predictions arise from prosodic and paralinguistic cues or from lexical shortcuts in transcripts. Current evaluations therefore leave open a central question: to what extent do LALMs genuinely listen rather than read? Addressing this gap, LISTEN systematically manipulates lexical–acoustic alignment to disentangle transcript reliance from acoustic sensitivity, offering the first diagnostic framework for assessing whether LALMs truly process emotional information from speech.
\paragraph{Large Audio Language Models}
Recent advances in large audio language models (LALMs) such as GPT-4o \cite{openai2024gpt4ocard}, Gemini 2.5 \cite{comanici2025gemini25pushingfrontier}, Qwen 2.5-Omni \cite{xu2025qwen25omnitechnicalreport}, Qwen 3-Omni \cite{xu2025qwen3omnitechnicalreport}, and Baichuan-Omni \cite{li2025baichuanomni15technicalreport} have demonstrated strong capabilities in processing spoken input and generating naturalistic responses. Baichuan-Omni and the Qwen Omni series are explicitly reported to derive from text-based LLM backbones later adapted to audio through multimodal training. While this design transfers rich linguistic knowledge from text, it also risks introducing a transcript-dominant bias in which lexical information overrides acoustic evidence. Gemini 2.5, though highly multimodal, has not disclosed its training specifics, leaving similar concerns unresolved. 
\paragraph{Speech Emotion Recognition}
Recent advances in deep learning have markedly improved Speech Emotion Recognition (SER). End-to-end CNN–RNN architectures can learn hierarchical acoustic representations from waveforms or spectrograms, achieving high accuracy and robustness across speakers and recording conditions \cite{barhoumi2024real}. More recent transformer and attention-based systems further enhance performance through intra-speech feature fusion, and graph-based modeling \cite{singh2023speech,chowdhury2025speech}. These approaches show that explicitly modeling prosodic, spectral, and temporal features yields strong emotion recognition accuracy from speech alone. Nevertheless, most SER systems remain confined to the acoustic modality and do not assess how lexical and acoustic cues jointly contribute to emotion understanding.



\section{The LISTEN Framework}

\label{sec:overview}
LISTEN is an evaluation framework designed to assess how large audio language models balance reliance on lexical cues versus acoustic cues in speech emotion understanding. The primary goal of LISTEN is not simply to measure classification accuracy, but to provide a controlled setting that disentangles lexical dependence from acoustic sensitivity, thereby enabling deeper insight into whether models are genuinely \emph{listening} to speech.

The benchmark is organized into four carefully designed conditions that manipulate the alignment between lexical cues and acoustic cues:  
(i) \textbf{Neutral-Text}, where lexical contents are emotionally neutral but acoustic cues vary, isolating the contribution of acoustic cues;  
(ii) \textbf{Emotion-Matched}, where lexical and acoustic cues are aligned to reinforce the same affect;  
(iii) \textbf{Emotion-Mismatched}, where lexical and acoustic cues conflict, as in sarcasm; and  
(iv) \textbf{Paralinguistic}, where affect is conveyed without lexical content (e.g., laugh, sighs, breathing).  
Within the first three conditions, LISTEN further compares performance across Text-only (lexical cues only), Audio-only (acoustic and implicit lexical cues), and Text+Audio (both modalities with explicit lexical reinforcement) modalities to probe modality-specific reliance.

This design is grounded in psycholinguistic findings that lexical and acoustic channels do not contribute equally in all contexts: when the two align, both provide useful evidence, but when they conflict, as in sarcasm, acoustic cue carries the decisive signal \cite{bryant2005there,attardo2003multimodal}. Our goal is not to require LALMs to mimic human processing, but to test whether they can adaptively exploit the information present in speech, identifying which cues are informative and when they should be weighted more heavily. In emotion-matched situations, lexical meaning may be useful, but in mismatched or paralinguistic settings, models must rely on acoustic and nonverbal signals to reach the correct interpretation. LISTEN makes these trade-offs explicit: by manipulating lexical–acoustic alignment, it forces models to reveal whether they are genuinely leveraging speech audio or defaulting to transcript shortcuts. 

\subsection{Data Construction}  
Selected samples from the LISTEN benchmark are shown in \autoref{fig:LISTEN_example}. Our benchmark construction follows a three-stage procedure to ensure theoretical grounding, dataset diversity, and quality control.  
\paragraph{Stage 1: Condition Design.}  
Building on the four experimental conditions, we formalize operational definitions and modality-specific ground-truth mapping to guide dataset selection and annotation. Neutral-Text: lexically neutral sentences with varied emotional acoustic cues; the text-only ground truth is fixed to neutral, while audio-only and text+audio use the sample’s true emotion label. Emotion-Matched: lexical meaning and acoustic cues are aligned; all three modalities use the true emotion label. Emotion-Mismatched: controlled conflicts between lexical polarity and prosodic tone; text-only uses the dataset’s explicit emotion label, whereas audio-only and text+audio use the dataset’s implicit emotion label. Paralinguistic: no lexical content; use the true emotion label. Representative examples appear in the Appendix \ref{app:condition_examples}.

\paragraph{Stage 2: Dataset Mapping.}
Each condition is instantiated using established emotional speech corpora spanning monologue and dialogue, acted and spontaneous data.
Neutral-Text includes acted corpora such as CREMA-D \cite{cao2014cremad}, Emotion Speech Dataset \citep{zhou2021emotional}, TESS \citep{schuller2010tess}, SAVEE \citep{king2011savee}, and RAVDESS \citep{livingstone2018ravdess}, which are specifically designed with fixed, semantically neutral sentences (e.g., “It’s eleven o’clock”) spoken with varied emotional prosody, ensuring affect is conveyed solely through acoustic cues. Emotion-Matched covers both acted and spontaneous corpora, including IEMOCAP \citep{busso2008iemocap}, CMU-MOSEI \citep{zadeh2018cmu}, OMGEmotionCh \citep{liu2021omgemotionchallenge}, MSP-PODCAST \citep{gladstone2020podcast}, and MELD \citep{chen2020mels}, where lexical semantics and prosody are broadly aligned;
Cue alignment for this group was verified through corpus-level metadata. Emotion-Mismatched leverages MUSTARD++ \citep{ray2022multimodal}, which contains sarcastic and ironic speech where lexical sentiment and prosodic emotion deliberately conflict; to verify this divergence, we manually examined the explicit emotion conveyed by lexical sentiment and the implicit emotion expressed through acoustic delivery in the corpus metadata. Finally, paralinguistic samples are extracted from IEMOCAP using annotated transcripts. Additional information on each dataset can be found in Appendix~\ref{app:data_intro}.

\paragraph{Stage 3: Question Generation.}
For each condition and modality (Text-only, Audio-only, Text+Audio), standardized multiple-choice emotion recognition prompts were generated using GPT-5 \cite{GPT5SystemCard}. Parallel templates targeted the same judgment, identifying the speaker’s emotion, while differing in focus on lexical meaning, vocal prosody, or both. To minimize inference-time bias, one paraphrased question form was randomly selected per item from a small pool of semantically equivalent variants, and answer options were shuffled to prevent positional bias. All prompt templates were manually reviewed to ensure semantic equivalence and modality relevance. Representative examples are provided in Appendix~\ref{app:qestion}.


\subsection{LISTEN Statistics}  
\autoref{tab:intra_corpus_results} summarizes the core statistics of the LISTEN benchmark, which consists of 7,939 evaluation questions spanning four experimental conditions and three modality partitions. Among the four conditions, \textit{Neutral-Text} constitutes the largest portion with 3,428 questions, followed by \textit{Emotion-Matched} (3,155), \textit{Paralinguistic} (975), and \textit{Emotion-Mismatched} (381). On average, each question contains 14.7 words, while answer options average 8.3 words. The associated audio clips are short and focused, averaging 3.5 seconds, ensuring that emotional cues remain perceptually salient without introducing long-context confounds. Detailed distributions for each dataset and condition are provided in the Appendix~\ref{app:distribution}.

\begin{table}[htbp]
    \centering
    \caption{Key statistics of the LISTEN benchmark.}
    \label{tab:intra_corpus_results}
    \begin{tabular}{l c}
    \toprule
    \textbf{Statistic} & \textbf{Value} \\
    \midrule
    Total questions & 7,939 \\
    Task count & 4 \\
    Modality count & 3 \\
    \midrule
    Neutral-Text & 3,428 \\
    Emotion-Matched & 3,155 \\
    Emotion-Mismatched & 381 \\
    Paralinguistic & 975 \\
    \midrule
    Average question length & 14.7 words \\
    Average option length & 8.3 words \\
    Average audio length & 3.5 seconds \\
    \bottomrule
    \end{tabular}
\end{table}

\section{Experiments}

\paragraph{Models}

We evaluate state-of-the-art large audio language models including closed-weight models, Gemini 2.5 Flash and Gemini 2.5 Pro \cite{comanici2025gemini25pushingfrontier}; open-weight models: Qwen2.5-Omni-7B \cite{xu2025qwen25omnitechnicalreport}, Qwen3-Omni-30B \cite{xu2025qwen3omnitechnicalreport}, Baichuan-Omni-1.5 \cite{li2025baichuanomni15technicalreport}, and Qwen3-Instruct \cite{xu2025qwen3omnitechnicalreport}. The hyperparameters and configurations used during the evaluation process are consistent with their official settings. Details appear in Appendix~\ref{app:model_detail}.

\paragraph{Evaluation Protocol}

For each sample, we present the model with a multiple-choice question containing 5-10 emotion options (happiness, sadness, anger, fear, surprise, disgust, neutral, frustration, excitement, ridicule). To prevent position bias, we randomize the order of choices for each query. Models receive only the audio, text, or both depending on the condition, with sample identifiers anonymized to prevent information leakage.

We employ zero-shot evaluation with carefully designed prompts that instruct models to classify emotions based solely on the provided input. For text-only conditions, models receive transcriptions without timing or acoustic annotations. For audio-only conditions, models process raw audio without text transcripts. For multimodal conditions, both audio and text transcripts are provided simultaneously. Examples are provided in Appendix~\ref{app:condition_examples}.

\subsection{Metrics}~\label{metrics}
We evaluate model performance using overall \textbf{accuracy}, the proportion of correctly classified samples across emotion categories. Since each sample is assigned a single ground-truth label, accuracy reflects the correctness and equals micro-F1 in this single-label multi-class setting.
To interpret model performance relative to chance, we report three reference baselines for each experiment:
\begin{itemize}
\item \textbf{Uniform Guess:} assumes a random classifier that predicts each emotion class with equal probability. This baseline represents the accuracy expected from purely random guessing with no prior information about the dataset.
\item \textbf{Majority Guess:} always predicts the most frequent emotion in the dataset. This provides an upper bound for trivial label-frequency heuristics and reflects dataset imbalance.
\item \textbf{Prediction-Marginal Distribution Baseline:} estimates the expected accuracy of a random classifier that samples predictions according to the model’s own empirical prediction distribution rather than uniformly. This captures how much of a model’s performance can be attributed to its output bias rather than meaningful input sensitivity.
\end{itemize}
Formally, let $p_i$ denote the probability the model predicts class $i$, and $q_i$ the empirical frequency of class $i$ in the ground-truth labels. The expected prediction-marginal accuracy is computed as:
\begin{equation}
\mathbb{E}[\text{Acc}] = \sum_i p_i q_i.
\end{equation}
This expectation accounts for both dataset imbalance ($q_i$) and model prediction bias ($p_i$), yielding a more informative lower bound than uniform guessing. When $p_i = q_i$, the expected value corresponds to the Bayes-optimal random baseline given the dataset label prior.
We report all accuracies in Table~\ref{tab:intra_corpus_results_detailed}, with each model’s prediction-marginal baseline shown in parentheses for direct comparison. The gaps between the accuracy reported and the prediction marginal distribution baseline reveal how much performance comes from the interpretation of input cues rather than from model's own prediction bias. Larger gaps indicate stronger use of real lexical or acoustic information.

\section{Results and Analysis}
\label{sec:results}
\begin{table*}
  [t]
  \centering
  \caption{Accuracy (with prediction marginal distribution baseline in parenthesis—see \autoref{metrics}) are reported. \textbf{Bold} values highlight the
highest value and \underline{underlined} values highlight the second-highest value in each experiment. For each condition, the Average column represents the mean of the audio and text+audio settings. 
The Overall Average is computed as the mean accuracy across all audio and text+audio results from the four experimental conditions (seven modalities in total).}
  \label{tab:intra_corpus_results_detailed}
  \large 
  \adjustbox{width=\textwidth,center}{ \begin{tabular}{l cccc cccc cccc cc}\toprule & \multicolumn{4}{c}{\textbf{Neutral-Text}} & \multicolumn{4}{c}{\textbf{Emotion-Matched}} & \multicolumn{4}{c}{\textbf{Emotion-Mismatched}} & \textbf{Paralinguistic} & \textbf{Overall} \\
  \cmidrule(lr){2-5} \cmidrule(lr){6-9} \cmidrule(lr){10-13} \cmidrule(lr){14-14} \cmidrule(lr){15-15} \textbf{Model} & Text & Audio & Text+Audio & Average & Text & Audio & Text+Audio & Average & Text & Audio & Text+Audio & Average & Audio & Average \\
  \midrule Uniform Guess & 12.5 & 12.5 & 12.5 & 12.5 & 6.7 & 6.7 & 6.7 & 6.7 & 10.0 & 10.0 & 10.0 & 10.0 & 12.5 & 10.1\\
  Majority Guess & 100 & 16.9 & 16.9 & 16.9 & 26.5 & 26.5 & 26.5 & 26.5 & 39.0 & 39.0 & 39.0 & 39.0 & 32.6 & 28.2\\
  \midrule\multicolumn{15}{c}{\textbf{Open-Weight Model}} \\
  \midrule Qwen3-instruct & 66.6 (65.0) & -- & -- & -- & 33.5 (12.4) & -- & -- & -- & \textbf{38.0} (29.8) & -- & -- & -- & -- & -- \\ 
  Qwen2.5-Omni-7B & \underline{85.4} (84.1) & \underline{34.0} (10.9) & 19.8 (11.8) & 26.9 & 36.4 (15.7) & 36.6 (14.1) & 38.6 (15.7) & 37.6 & 34.0 (13.5) & \textbf{38.5} (29.5) & \underline{39.1} (23.6) & \underline{38.8} & \textbf{22.7} (11.4) & 32.8\\
  Qwen3-Omni-30B & \underline{85.4} (84.0) & 29.3 (11.3) & \underline{25.3} (11.8) & \underline{27.3} & \underline{38.7} (11.7) & \textbf{42.4} (15.8) & \textbf{43.1} (15.5) & \textbf{42.8} & \underline{34.6} (12.2) & \underline{37.4} (26.7) & \underline{39.1} (25.2) & 38.3 & \underline{21.0} (12.6) & \underline{33.9}\\
  Baichuan-Omni-1.5 & 81.2 (79.6) & 16.5 (11.5) & 15.2 (12.1) & 15.9 & 31.0 (15.9) & 36.0 (15.6) & 36.0 (17.8) & 36.0 & 31.0 (27.5) & 36.0 (31.0) & 36.0 (32.0) & 36.0 & \textbf{22.7} (11.5) & 28.3\\
  \midrule\multicolumn{15}{c}{\textbf{Closed-Weight Model}} \\
  \midrule Gemini2.5-Flash & 82.5 (81.5) & 25.6 (10.7) & 24.6 (11.0) & 25.1 & 36.6 (14.3) & 30.7 (12.3) & 38.9 (14.1) & 34.8 & 33.1 (10.1) & 35.8 (15.5) & 38.0 (18.5) & 36.9 & 18.0 (12.0) & 30.2\\
  Gemini2.5-Pro & \textbf{96.6} (96.6) & \textbf{34.9} (12.8) & \textbf{41.7} (12.9) & \textbf{38.3} & \textbf{38.8} (15.7) & \underline{37.6} (16.9) & \underline{40.2} (14.3) & \underline{38.9} & 31.8 (10.0) & 36.9 (22.5) & \textbf{42.6} (23.3) & \textbf{39.8} & 15.7 (9.3) & \textbf{35.7}\\
  \bottomrule\end{tabular} }
\end{table*}
\begin{figure*}[h]
  \centering
  \includegraphics[height=0.18\textheight,width=\linewidth]{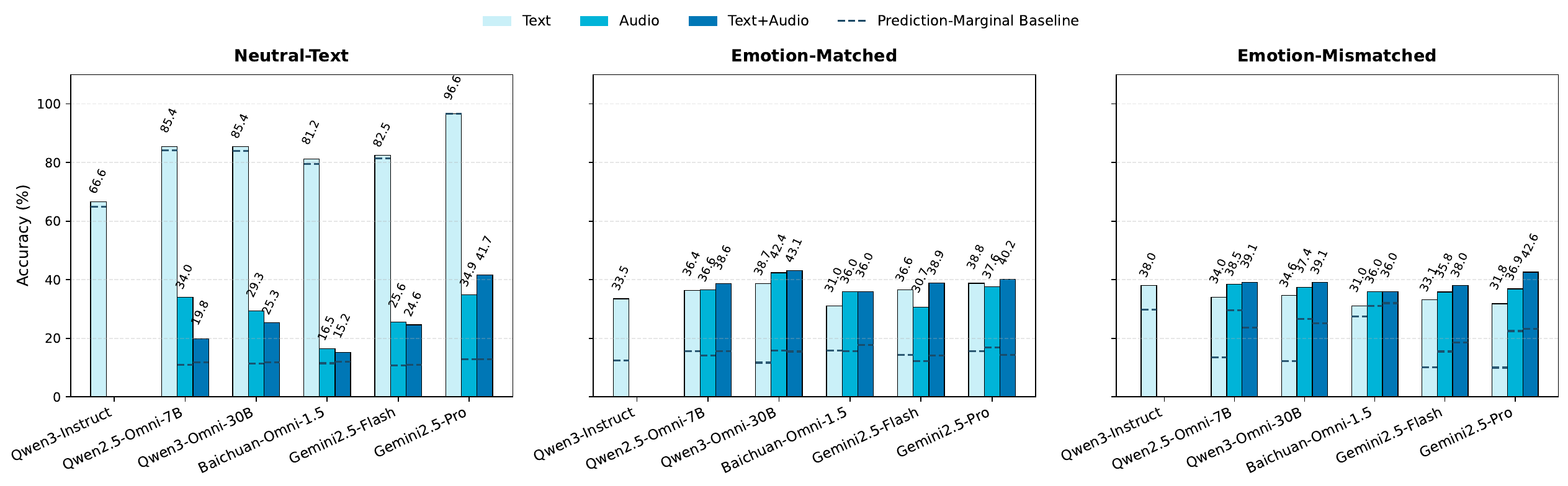}
 \caption{Model accuracy across three LISTEN conditions (Neutral-Text, Emotion-Matched, Emotion-Mismatched) under text-only, audio-only, and text+audio modalities. Dashed lines indicate prediction-marginal baselines.}
  \label{fig:bar}
\end{figure*}
\begin{figure*}[h]
  \centering
  \includegraphics[height=0.4\textheight, width=\linewidth]{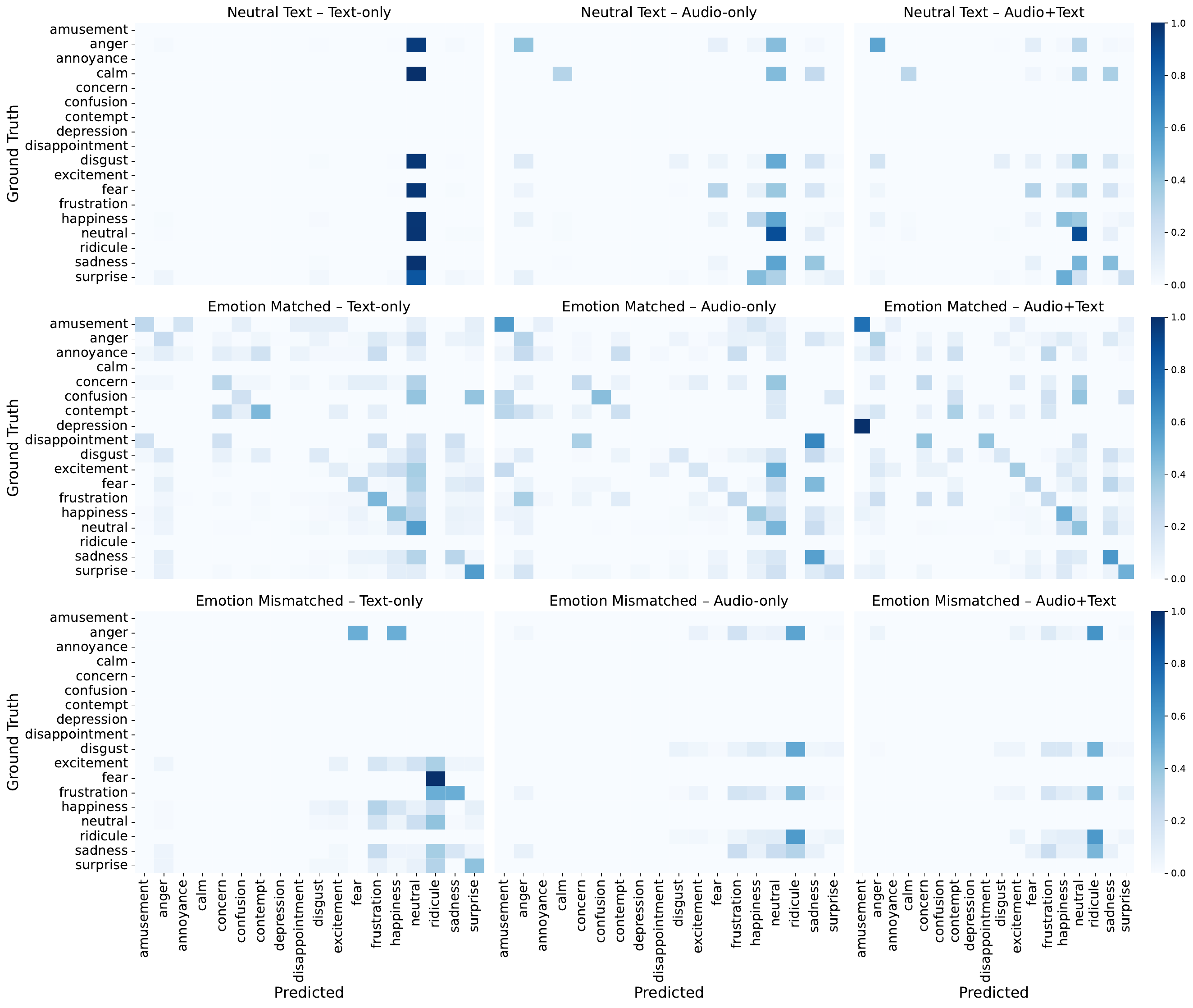}
  \caption{Confusion matrices showing Gemini 2.5 Pro's emotion recognition performance across three experimental conditions in the LISTEN benchmark. Row-normalized matrices display prediction distributions for each true emotion class across: (1) Neutral Text, (2) Emotion Matched, and (3) Emotion Mismatched conditions, each tested with text-only, audio-only, and audio+text modalities.}
  \label{fig:confusion_matrices}
\end{figure*}
\begin{figure}[h]
  \centering
  \includegraphics[ width=\linewidth]{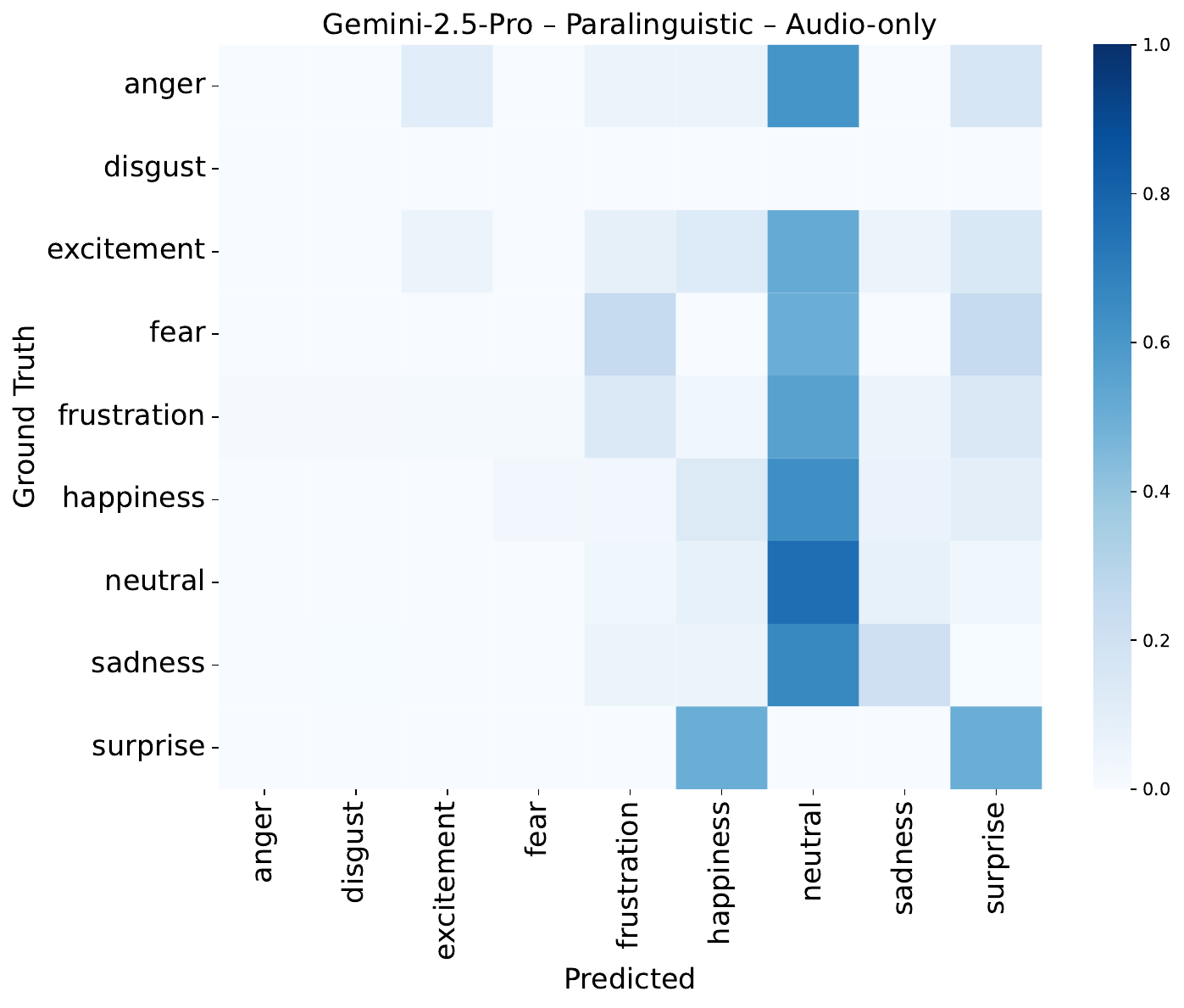}
  \caption{Confusion matrices showing Gemini 2.5 Pro's emotion recognition performance in paralinguistic condition}
  \label{fig:confusion_matrices_para}
\end{figure}

\paragraph{Overall Performance}

\autoref{tab:intra_corpus_results_detailed} and \autoref{fig:bar} summarizes results across all experimental conditions. Overall, models still rely more on text than on acoustic cues, but their behavior shifts depending on how speech and lexical cues align.

\paragraph{Neutral-Text condition} the transcripts are deliberately emotionless while the audio expresses a range of emotions, allowing us to isolate the role of acoustic cues. In the text-only setting, where all transcripts are neutral and the ground-truth label is neutral for every sample, LALMs reach near-perfect accuracy (e.g., 96.6\% for Gemini2.5-Pro, 85.4\% for the Qwen Omnis). In contrast, in audio-only setting, models must infer emotion directly from acoustic cues. It shows much lower scores (25–35\%), revealing the difficulty of recognizing emotional tone from acoustic cues. Qwen2.5-Omni-7B and Gemini 2.5-Pro has 34.0\% and 34.9\% accuracy relatively. Other models' accuracy are all below 30.0\% When text and audio are combined, the closed-weight Gemini2.5-Pro shows a modest gain (41.7\%), suggesting partial use of acoustic cues even when lexical content is neutral, whereas open-weight models (e.g., Qwen2.5-Omni-7B, Baichuan-Omni-1.5) and Gemini2.5-Flash often perform worse than their audio-only baselines. Qwen2.5-Omni-7B has a large accuracy drop from audio only to text+audio (34.0\% → 19.8\%)

Further evidence of this pattern is shown in the confusion matrices in \autoref{fig:confusion_matrices}, which visualizes Gemini 2.5-Pro’s emotion-recognition behavior across the three LISTEN conditions—(1) Neutral-Text, (2) Emotion-Matched, and (3) Emotion-Mismatched—each tested with text-only, audio-only, and audio + text inputs. The top row three heat maps corresponds to the Neutral-Text condition. In the text-only setting, the confusion matrix shows a single dominant bar in the neutral column, indicating that the model predicts “neutral” for nearly all samples. In both the audio-only and audio + text settings, a faint diagonal emerges, showing that the model can identify a subset of emotional classes from acoustic cues. However, the same vertical concentration in the neutral column persists, showing that the neutral lexical content biases the model’s interpretation even when the speech conveys clear emotional tone. Overall, this pattern indicates that LALMs over-emphasize lexical cues, leading to interference rather than effective identify emotion based on acoustic cues. More details are shown in \autoref{fig:all_exp1}.

\paragraph{Emotion-Matched condition.}
In this setting, the speech and transcripts express the same emotion, allowing us to evaluate how LALMs process consistent lexical–acoustic alignment. Performance across models is relatively balanced between text and audio inputs, with modest multimodal gains suggesting limited but consistent cue integration. Among open-weight systems, Qwen3-Omni-30B achieves the highest scores: 38. 7\% (text only), 42. 4\% (audio only) and 43. 1\% (text + audio), showing that it effectively leverages both lexical and acoustic cues when they match. Qwen2.5-Omni-7B follows a similar pattern (36.4–38.6\%), while Baichuan-Omni-1.5 performs consistently lower (31–36\%), suggesting a more limited capacity for recognizing emotions from speech.

For closed-weight models, Gemini 2.5-Pro performs strongly and consistently across modalities: 38.8\% (text-only), 37. 6\% (audio-only) and 40.2\% (text + audio) demonstrating stable use of lexical and acoustic cues.
In contrast, Gemini 2.5-Flash performs worse when using only audio: its accuracy drops from 36.6\% with text to 30.7\% with audio, showing that it struggles to capture emotional tone even when acoustic cues and lexical cues have matched emotion in the audio. When both inputs are combined, its score rises again to 38.9\%, suggesting that adding text helps the model recover accuracy by providing a more stable signal.

These trends are illustrated in \autoref{fig:confusion_matrices}, where the three matrices in the second row (Emotion-Matched condition) for Gemini 2.5 Pro display a faint diagonal.In the text-only and audio-only settings, a visible neutral column remains, indicating some bias toward neutral predictions. In text+audio setting, the diagonal becomes slightly clearer and the neutral column weakens, indicating that emphasizing lexical content enhances discriminability across emotion categories when textual and acoustic signals are aligned. Overall, although most LALMs can recognize emotion better when lexical and acoustic cues point to the same label, compared to neutral-text, lexical emotion recognition (text-only) is still weak, and combining lexical and acoustic acoustic emotion recognition (audio-only, text+audio) yields little improvement. Some architectures (like Gemini 2.5-Flash) still struggle to exploit acoustic information effectively. Confusion matrices for all models are shown in \autoref{fig:all_exp2}.

\paragraph{Emotion-Mismatched condition.}
In this condition, the speech and text express opposing emotions, testing how LALMs weight lexical vs. acoustic cues when they conflict. The results suggest that LALMs can detect sarcasm when lexical and acoustic cues disagree but struggle to identify the specific sarcastic emotion underlying that mismatch.

Among open-weight models, both Qwen2.5-Omni-7B (34.0\% → 38.5\% → 39.1\%) and Qwen3-Omni-30B (34.6\% → 37.4\% → 39.1\%) gain modest improvements from audio. Baichuan-Omni-1.5 (31.0\% → 36.0\% → 36.0\%) has lower accuracy across all three settings, and the difference between these accuracy and the prediction marginal distribution baseline is small, suggesting limited discriminative use of input signals. For closed-weight models, both Gemini2.5-Flash (33.1\% → 35.8\% → 38.0\%) and Gemini2.5-Pro (31.8\% → 36.9\% → 42.6\%) show clearer acoustic utilization. Gemini2.5-Pro achieves 42.6\% in the text+audio setting versus a 23.3\% prediction-marginal baseline, demonstrating better sensitivity to acoustic cues under cue conflict.

However, the overall gaps between reported accuracies and prediction marginal distribution baseline are smaller here than in the Neutral-Text or Emotion-Matched conditions, implying that much of the observed performance may come from prediction biases toward a few dominant emotional categories rather than fine-grained emotion recognition in lexical and acoustic cues. The heat maps in \autoref{fig:confusion_matrices} support this interpretation: in the Emotion-Mismatched condition (bottom row), the text-only setting shows weak diagonals concentrated around a few emotions, mainly happiness, neutral, surprise, and sadness. However, both the audio-only and text+audio settings show strong vertical bars on ridicule and, to a lesser extent, frustration, indicating that models classify a large portion of conflicting samples into these categories. This pattern suggests that LALMs recognize the presence of emotional conflict in lexical and acoustic cues but resolve it by collapsing diverse sarcastic emotions (anger, disgust, frustration, ridicule, sadness) into only two classes: ridicule and frustration,  revealing a key limitation in current LALM understanding of complex sarcastic emotions. We observe similar performance in Qwen2.5-omni, Qwen3-omni, Baichuan-omni, and Gemini2.5-Flash. More details are shown in \autoref{fig:all_ex3}.

\paragraph{Paralinguistic condition.}
This condition isolates nonlexical affective cues, such as laughter, sighs, gasps, or other vocalizations, by removing linguistic content entirely. Performance across models are just above uniform guess baseline and prediction marginal distribution baseline, showing that current LALMs have limited ability to interpret emotion from nonverbal sounds alone. Among open-weight systems, Qwen2.5-Omni-7B and Baichuan-Omni-1.5 reach the highest accuracy (22.7\%), followed by Qwen3-Omni-30B (21.0\%), while the closed-weight Gemini2.5-Flash (18.0\%) and Gemini2.5-Pro (15.7\%) perform slightly lower. The confusion matrix for Gemini2.5-Pro in Figure~\ref{fig:confusion_matrices} illustrates this limitation. A faint diagonal suggests partial recognition of emotions like surprise and sadness, while the dominant neutral column reveals a strong bias toward predicting “neutral” for nearly all categories except "surprise". This bias shows limited ability to distinguish nonverbal emotions and reliance on lexical cues for emotion identification. See Appendix~\ref{app:radar}, for detailed model results.

\section{Discussion and Conclusion}

\paragraph{Where Do Models Succeed and Fail? Lexical Dominance and the Limits of Listening.}
LISTEN reveals a mixed and fragile affective skill profile in current LALMs. In the Neutral-Text, Audio-only, and Text+Audio conditions, models remain strongly biased toward predicting “neutral,” reflecting an overreliance on lexical cues and a tendency to default to neutral interpretations regardless of acoustic cues. In the Paralinguistic condition, where no lexical content is available at all, models still default to “neutral,” not because of lexical interference, but because of the absence of lexical grounding. This pattern suggests that current LALMs depend on textual information both as an interpretive anchor and as a confidence cue: when lexical guidance is strong, they ignore acoustic variation, and when it is missing, they revert to neutral as the safest default. 

When lexical and acoustic cues align, as in the Emotion-Matched condition, models can recognize a wider range of emotions, but overall accuracy remains low and the neutral bias persists. This suggests that their emotion recognition benefits only marginally from cue consistency and that effective multimodal integration is still limited in LALMs.

When lexical and acoustic cues conflict, as in the Emotion-Mismatched condition, LALMs can detect that an emotional discrepancy exists but fail to interpret it precisely. They can identify such conflicts as sarcasm, collapsing diverse sarcastic emotional states (e.g., anger, disgust, or ridicule) into a narrow set of categories. This indicates that while models sense the presence of incongruity between modalities, they still lack the ability to reason about the nuanced emotional or pragmatic meaning behind that conflict. 

\paragraph{Implications and Future Directions.}
In general, LISTEN highlights a central finding behind our question: Audio LLMs often transcribe more than they truly listen. While they can detect emotional variation in speech, their interpretations remain shallow and heavily guided by lexical information. Even when provided with rich acoustic cues, models tend to default to “neutral” predictions, revealing limited prosodic sensitivity and weak integration between text and audio streams.
Future progress in “listening” models may benefit from the advances achieved in corporate and state-of-the-art speech emotion recognition (SER) systems, which attain strong accuracy by explicitly modeling prosodic, spectral, and temporal features of speech. Techniques on emotional speech could be adapted to improve prosodic grounding and acoustic sensitivity in LALMs. Building on these insights, next-generation audio language models should not only perceive acoustic variation but also infer its emotional and communicative intent.

\section*{Limitations}
While LISTEN prioritizes controlled and interpretable evaluation of lexical–acoustic cue reliance in LALMs, several scope boundaries are worth noting.
First, each utterance is presented independently rather than in a full conversational context, allowing precise manipulation of the lexical-acoustic alignment but limiting access to broader pragmatic cues. This design isolates local emotional expression, though future extensions may reintroduce discourse-level context to examine contextual modulation of affect.
Second, the benchmark centers on emotion understanding, a core yet specific aspect of social reasoning, which can be expanded to other pragmatic and interactive dimensions.
These choices reflect intentional simplifications to enable diagnostic clarity, forming a controlled basis for future, context-rich multimodal benchmarks.
\section*{Ethical Considerations}
All datasets used in this work are publicly available and were originally collected under ethical or open-use guidelines. LISTEN is designed solely for research evaluation and does not involve new human data collection. We acknowledge that emotion recognition technologies carry potential privacy and misuse risks, and encourage future work to prioritize transparency, fairness, and responsible deployment.

We are thankful for the generous support of computational resources provided by the Ohio Supercomputer Center.

\bibliography{custom}
\newpage
\appendix

\section{Appendix}

\subsection{Source Datasets Used During Data Construction}
\label{app:data_intro}
Each experimental condition in our benchmark is instantiated using established emotional speech corpora spanning acted and spontaneous data, monologue and dialogue, and both congruent and incongruent emotion–text alignments.

\paragraph{CREMA-D} \cite{cao2014cremad}: The Crowd-Sourced Emotional Multimodal Actors Dataset consists of 7,442 clips from 91 actors (48 male, 43 female) portraying six emotions—anger, disgust, fear, happy, neutral, and sad—across 12 fixed sentences. Acted speech. The CREMA-D dataset is licensed under the Open Database License (ODbL v1.0) (Open Data Commons). This license permits sharing, use, and adaptation, but requires attribution and that derivative databases remain under the same license.

\paragraph{Emotion Speech Dataset} \cite{zhou2021emotional}: This dataset contains 350 parallel utterances spoken by 10 native Mandarin speakers, and 10 English speakers with 5 emotional states (neutral, happy, angry, sad and surprise) We only use English data. It provides consistent lexical content for analyzing prosodic variations in acted emotional speech. The RAVDESS is released under a Creative Commons Attribution-NonCommercial-ShareAlike 4.0 International License, CC BY-NC-SA 4.0

\paragraph{TESS} \cite{schuller2010tess}: The Toronto Emotional Speech Set contains 2,800 recordings of two female actors simulating seven emotions—anger, disgust, fear, happiness, pleasant surprise, sadness, and neutral—based on the same lexical template “Say the word \_\_\_.” It is designed for perceptual studies of emotional prosody. Attribution-NonCommercial-NoDerivatives 4.0 International (CC BY-NC-ND 4.0)

\paragraph{SAVEE} \cite{king2011savee}: The Surrey Audio-Visual Expressed Emotion dataset features 480 utterances from four male native English speakers—DC, JE, JK, and KL—who were postgraduate students at the University of Surrey, aged between 27 and 31. The dataset encompasses seven emotional classes: anger, disgust, fear, happiness, sadness, surprise, and neutrality, as commonly defined in psychological studies. Each speaker articulated 15 sentences per emotion, selected from the TIMIT corpus. This set included three sentences shared across all emotions, two that were emotion-specific, and ten general utterances tailored to each emotion, arranged in alphabetical order. License: Data files @ Original Authors 

\paragraph{RAVDESS} \cite{livingstone2018ravdess}: The Ryerson Audio-Visual Database of Emotional Speech and Song includes 7,356 audio and video clips from 24 professional actors expressing emotions through both speech and song. It encompasses calm, happy, sad, angry, fearful, surprise, and disgust, supporting multimodal emotion recognition. Licensed under CC BY-NA-SC 4.0.

\paragraph{IEMOCAP} \cite{busso2008iemocap}: The Interactive Emotional Dyadic Motion Capture Database comprises approximately 12 hours of audiovisual recordings from ten actors performing scripted and improvised dialogues. It is annotated for categorical emotions (anger, happiness, sadness, neutral, excitement) and dimensional affect ratings (valence, arousal, dominance), enabling analysis of spontaneous emotional dynamics. License: \url{https://sail.usc.edu/iemocap/Data_Release_Form_IEMOCAP.pdf}

\paragraph{CMU-MOSEI} \cite{zadeh2018cmu}: The Multimodal Opinion Sentiment and Emotion Intensity dataset contains 23,454 sentence-level video segments from online monologues annotated for both sentiment and emotion. It provides large-scale multimodal coverage of spontaneous speech in natural contexts. License: CC BY-SA 4.0 All data copyright: Carnegie Mellon University \& authors

\paragraph{OMG-Emotion Challenge Dataset}\cite{liu2021omgemotionchallenge}: This dataset includes monologue videos annotated for continuous emotion dimensions (valence and arousal), focusing on gradual emotion evolution within a single speaker. It emphasizes contextual and temporal modeling of affective expression. License: Apache License 2.0

\paragraph{MSP-Podcast} \cite{gladstone2020podcast}: A large-scale corpus of natural English speech extracted from public podcasts, containing over 100,000 segments annotated for categorical emotions and continuous dimensions. It captures rich acoustic variability and spontaneous emotional speech. This dataset has Common Licenses that permit the distribution of the corpus.

\paragraph{MELD} \cite{chen2020mels}: The Multimodal EmotionLines Dataset extends the EmotionLines corpus with audio and visual modalities from the TV series Friends. It includes 13,000 utterances from 1,433 dialogues annotated with seven emotions: anger, disgust, sadness, joy, neutral, surprise, and fear. GNU General Public License v3.0

\paragraph{MUSTARD++} \cite{ray2022multimodal}: The Multimodal Sarcasm Detection dataset extends the original MUSTARD with additional sarcastic and non-sarcastic clips from TV shows. It contains audio, visual, and textual modalities annotated for sarcasm, where lexical and prosodic cues intentionally conflict, making it ideal for evaluating multimodal incongruence. The data provides both explicit emotion annotation, aligned with lexical content, and implicit emotion annotation aligned with speech. License. CC0: Public Domain.

\subsection{Question Prompts}  
\label{app:qestion}
To ensure consistent evaluation across modalities, we design parallel sets of natural-language prompts for Text-only, Audio-only, and Text+Audio conditions. Below are representative examples for each modality.  

\paragraph{Text-only}  
\begin{itemize}
    \item Based on the content of this text, what emotion would the person likely be feeling?  
    \item What emotion is conveyed by the words in this statement?  
    \item Reading this text, what emotional state does the speaker appear to be in?  
    \item From the semantic content alone, what emotion is being expressed?  
    \item What feeling is suggested by the meaning of these words?  
    \item Based solely on the text content, what emotion would you identify?  
    \item What emotional tone is conveyed by the literal meaning of this statement?  
\end{itemize}  

\paragraph{Audio-only}  
\begin{itemize}
    \item What emotion is expressed in the speaker's voice?  
    \item What emotion does the speaker convey through their tone?  
    \item Based on the vocal expression, what emotion is the speaker feeling?  
    \item What emotional state is reflected in the speaker's voice?  
    \item How would you describe the emotional tone of the speaker?  
    \item What emotion is communicated through the speaker's vocal prosody?  
    \item Listening to the voice, what emotion is being expressed?  
\end{itemize}  

\paragraph{Text + Audio}  
\begin{itemize}
    \item Considering both the words and how they are spoken, what is the speaker's true emotional state?  
    \item What emotion is conveyed when you combine the text content with the vocal expression?  
    \item Based on both the semantic meaning and prosodic cues, what emotion is the speaker feeling?  
    \item How do the words and vocal tone together reveal the speaker's emotional state?  
    \item What emotion emerges when you integrate the textual and acoustic information?  
    \item Taking into account both content and delivery, what emotion is being expressed?  
    \item What is the complete emotional picture when combining words and voice?  
\end{itemize}

\subsection{LISTEN Emotion Distribution across experimental conditions}
\label{app:distribution}

\begin{figure*}[t]
  \centering
  \includegraphics[ width=\linewidth]{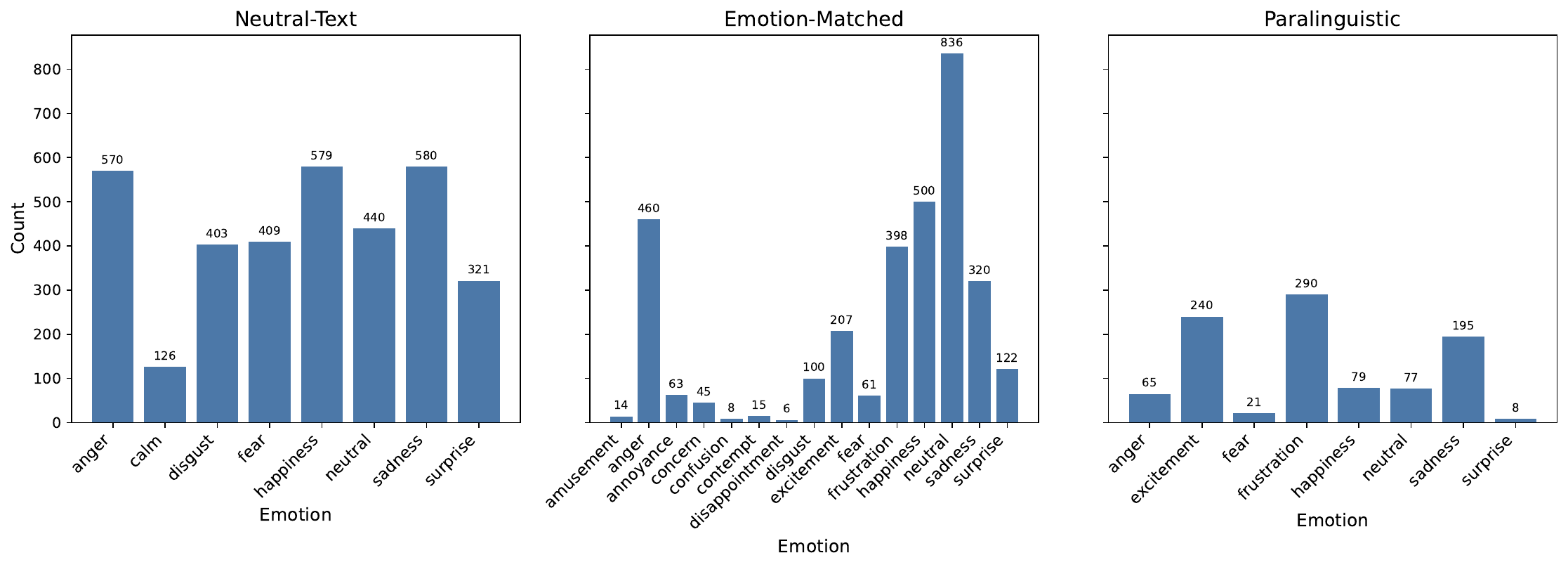}
  \caption{Ground-truth label distributions for Neutral-Text, Emotion-Matched, and Paralinguistic conditions. Each subplot shows counts for the labels present in that dataset.}
  \label{fig:emo_distri_main}
\end{figure*}

We visualize the distribution of ground-truth emotion labels for each experimental condition.
\paragraph{Neutral-Text, Emotion-Matched, and Paralinguistic}
\autoref{fig:emo_distri_main} aggregates three subplots, each showing per-dataset label counts using that dataset’s own taxonomy (after consistent normalization).
X-axes list only the emotions present in each dataset; the shared Y-axis facilitates cross-condition comparison of prevalence.

\begin{figure*}[t]
  \centering
  \includegraphics[ width=\linewidth]{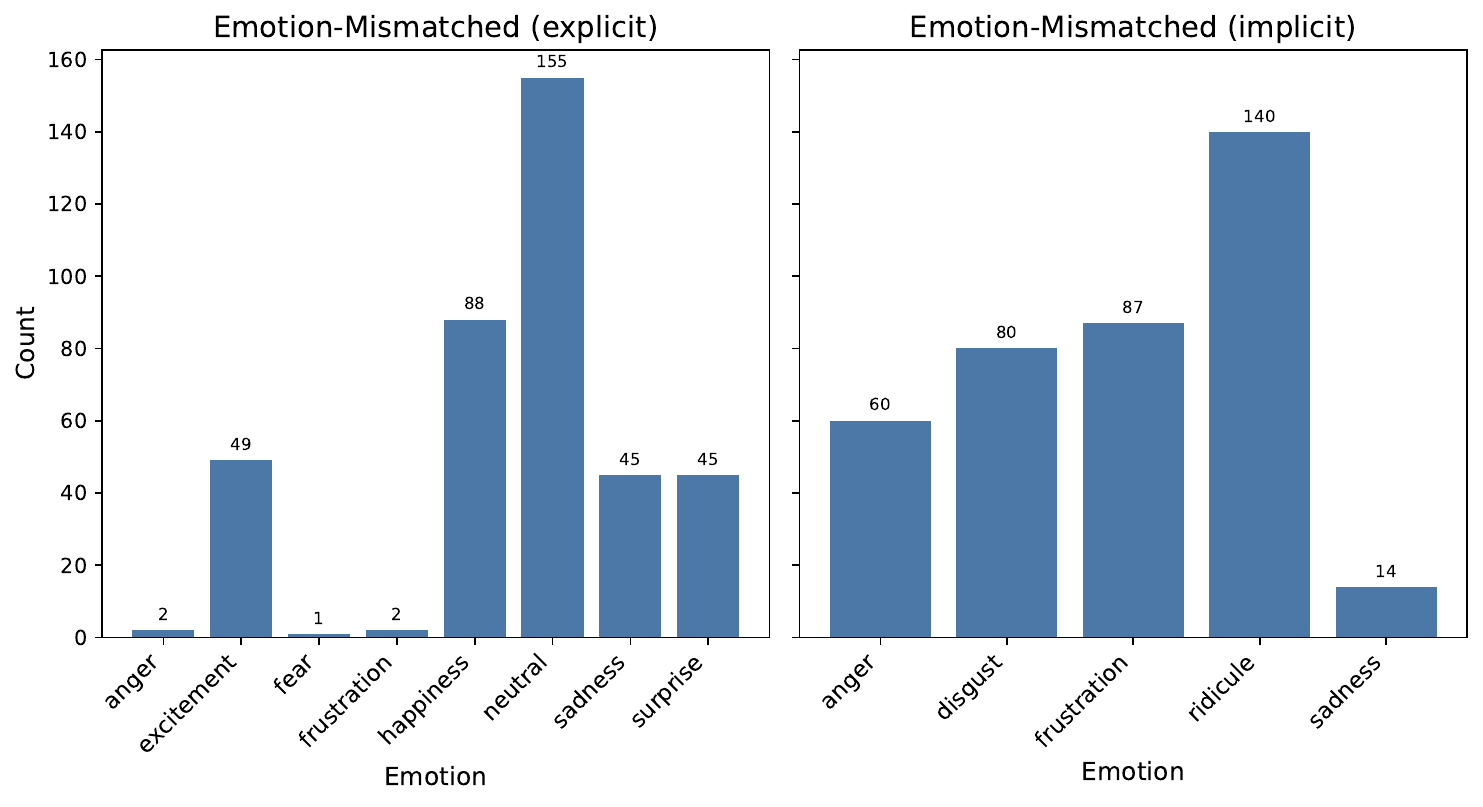}
  \caption{Ground-truth label distributions for Emotion-Mismatched (explicit) and Emotion-Mismatched (implicit).}
  \label{fig:emo_distri_mismatch}
\end{figure*}

\paragraph{Emotion-Mismatched (explicit) and Emotion-Mismatched (implicit)}
\autoref{fig:emo_distri_mismatch} isolates the two mismatched conditions, highlighting how label prevalence differs between explicit and implicit mismatches.
\newpage
\subsection{Large Audio Language Models Evaluated in LISTEN}
\label{app:model_detail}
We evaluate six recent large audio–language models (LALMs) spanning both open-weight and proprietary systems, as summarized in \autoref{tab:models}. 
The open-weight group includes the Qwen and Baichuan series, which represent the strongest publicly released multilingual models with unified speech–text understanding capabilities. 
All models are evaluated in zero-shot settings without fine-tuning to ensure comparability across conditions.
\begin{table*}[t]
\centering
\small
\caption{Large audio–language models evaluated in the LISTEN benchmark. “?” indicates unspecified public information.}
\label{tab:models}
\begin{tabularx}{\textwidth}{l X l l l l}
\toprule
\textbf{Model} & \textbf{Identifier} & \textbf{Creator} & \textbf{Access Type} & \textbf{Release Date} & \textbf{Params} \\
\midrule
\multicolumn{6}{l}{\textbf{Open-weight Audio–Language Models}} \\
\midrule
Qwen3-Instruct & Qwen3-4B-Instruct-2507 & Alibaba Cloud & Open-weight & 2025-08-06 & 4.02B \\
Qwen2.5-Omni (7B) & qwen2.5-omni-7b & Alibaba Cloud & Open-weight & 2025-05-13 & 10.7B \\
Qwen3-Omni (30B) & qwen3-omni-30b & Alibaba Cloud & Open-weight & 2025-09-026 & 35.3B \\
Baichuan-Omni (1.5) & baichuan-omni-1.5 & Baichuan Inc. & Open-weight & 2025-01-26 & 11B \\
\midrule
\multicolumn{6}{l}{\textbf{Closed-weight Audio–Language Models}} \\
\midrule
Gemini 2.5 Flash & gemini-2.5-flash & Google DeepMind & API & 2025-06 & ? \\
Gemini 2.5 Pro & gemini-2.5-pro & Google DeepMind & API & 2025-06 & ? \\
\bottomrule
\end{tabularx}
\end{table*}

\subsection{Detailed Cross-Modality Performance Analysis}
\label{app:radar}
\begin{figure*}[t]
  \centering
  \includegraphics[width=\linewidth]{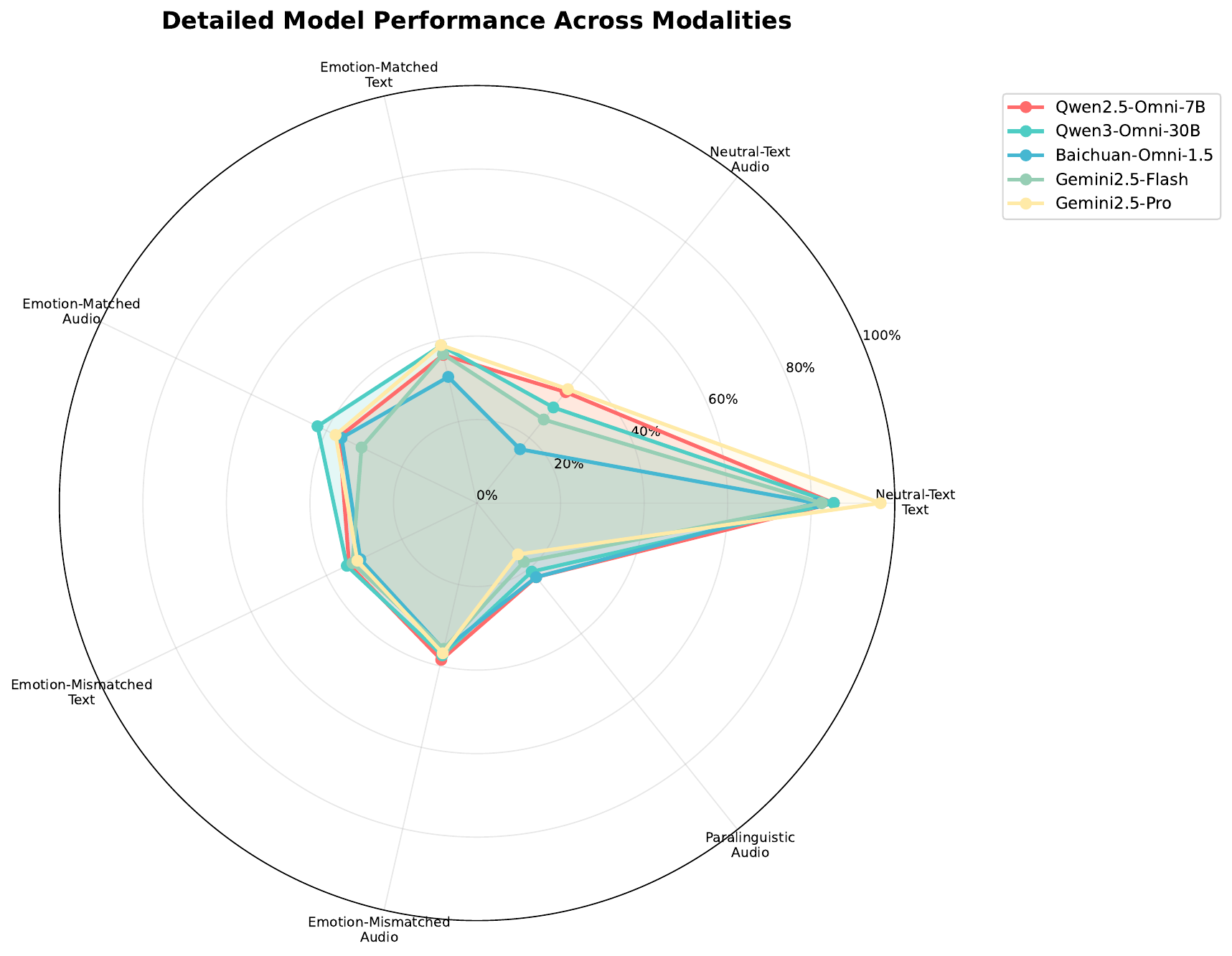}
  \caption{Radar plot comparing detailed model performance across all modality–condition pairs in the LISTEN benchmark. Each axis represents a specific evaluation setting (e.g., Neutral-Text, Emotion-Matched, Emotion-Mismatched, Paralinguistic) under text-only, audio-only, and text+audio modalities. }
  \label{fig:radar}
\end{figure*}

\autoref{fig:radar} compares detailed model performance across all modality–condition pairs in the LISTEN benchmark. Each axis represents a distinct evaluation setting, spanning Neutral-Text, Emotion-Matched, Emotion-Mismatched, and Paralinguistic conditions in both text and audio modalities.
Across models, accuracy peaks in the Neutral-Text condition, where lexical content alone provides strong cues for emotion inference. Performance declines sharply in Audio-only and Paralinguistic settings, confirming that current large audio–language models (LALMs) struggle to extract affective meaning from acoustic cues. Even in Emotion-Matched scenarios, where lexical and prosodic signals align, the gains remain modest, suggesting limited multimodal integration.
Among the evaluated systems, Gemini 2.5 Pro achieves the highest overall performance, followed by Qwen3-Omni-30B, yet all models display the same qualitative trend of lexical dominance. 
\newpage
\subsection{Confusion Matrices For All Models}
\label{app:confusion_matrics_all}
\begin{figure*}[t]
  \centering
  \includegraphics[ width=\linewidth]{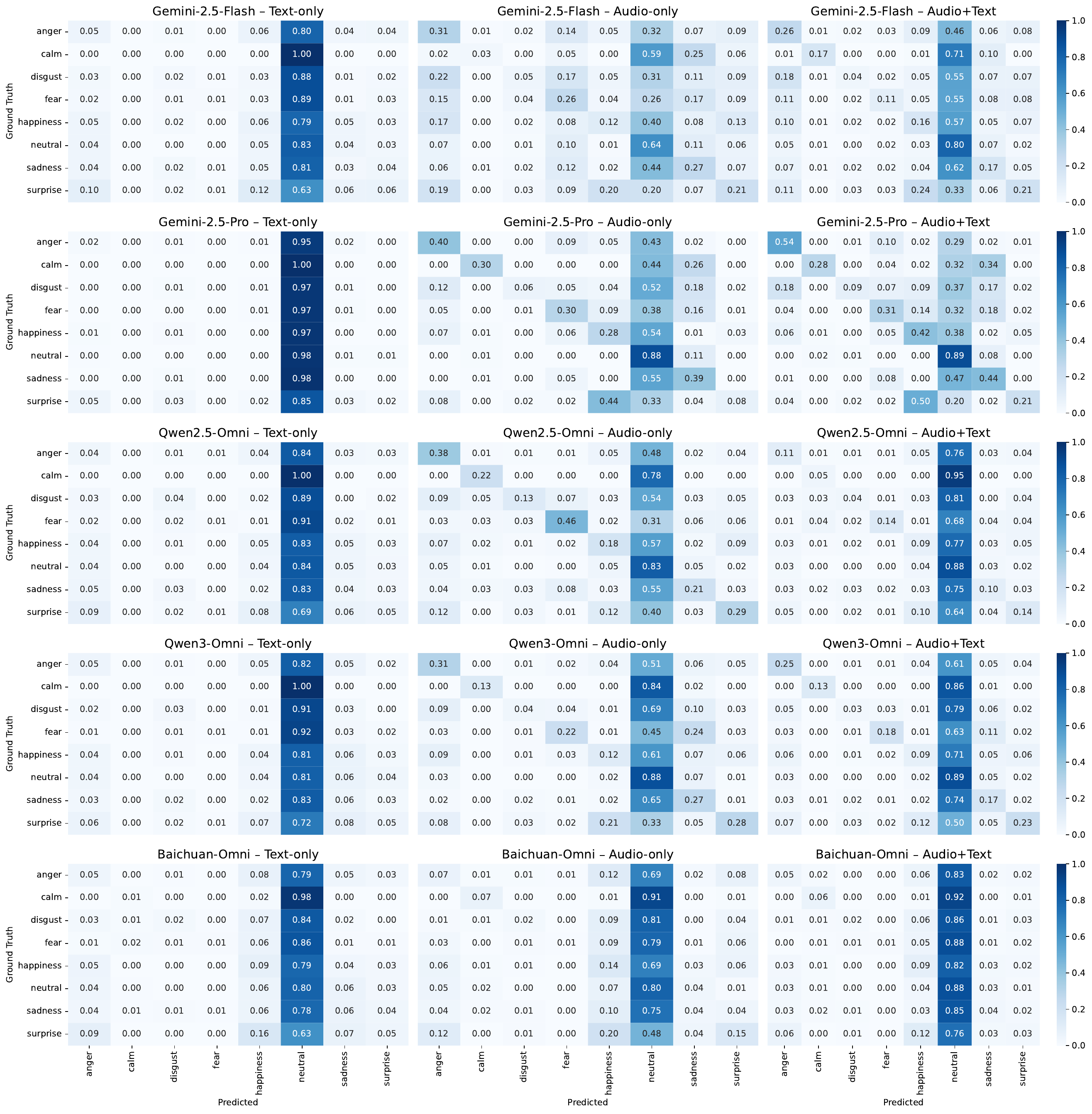}
  \caption{Confusion matrices showing all models' emotion recognition performance for Neutral-Text condition in the LISTEN benchmark. Row-normalized matrices display prediction distributions for each tested with text-only, audio-only, and audio+text modalities.}
  \label{fig:all_exp1}
\end{figure*}

\begin{figure*}[t]
  \centering
  \includegraphics[ width=\linewidth]{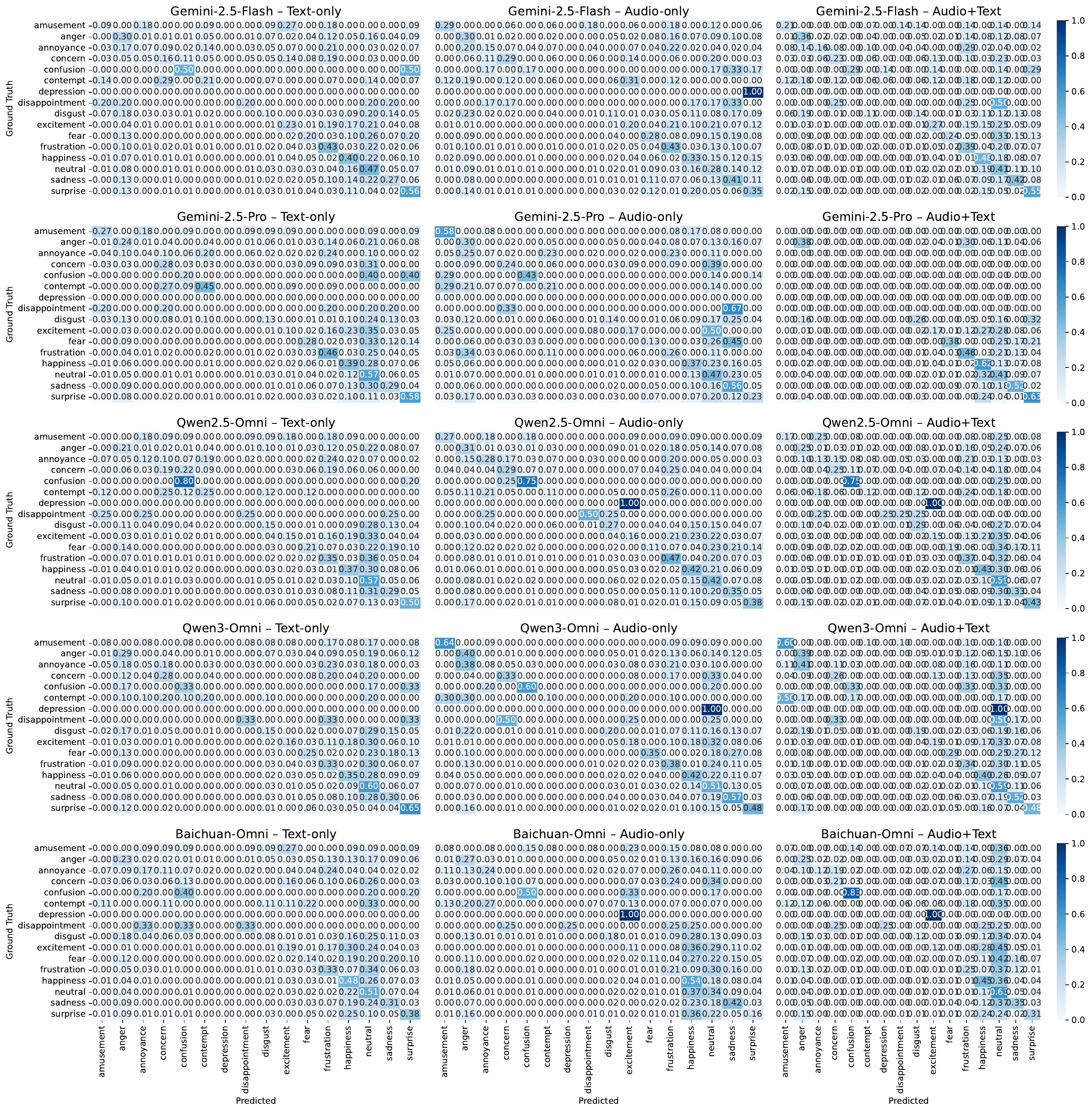}
  \caption{Confusion matrices showing all models' emotion recognition performance for Emotion-Matched condition in the LISTEN benchmark. Row-normalized matrices display prediction distributions for each tested with text-only, audio-only, and audio+text modalities.}
  \label{fig:all_exp2}
\end{figure*}

\begin{figure*}[t]
  \centering
  \includegraphics[ width=\linewidth]{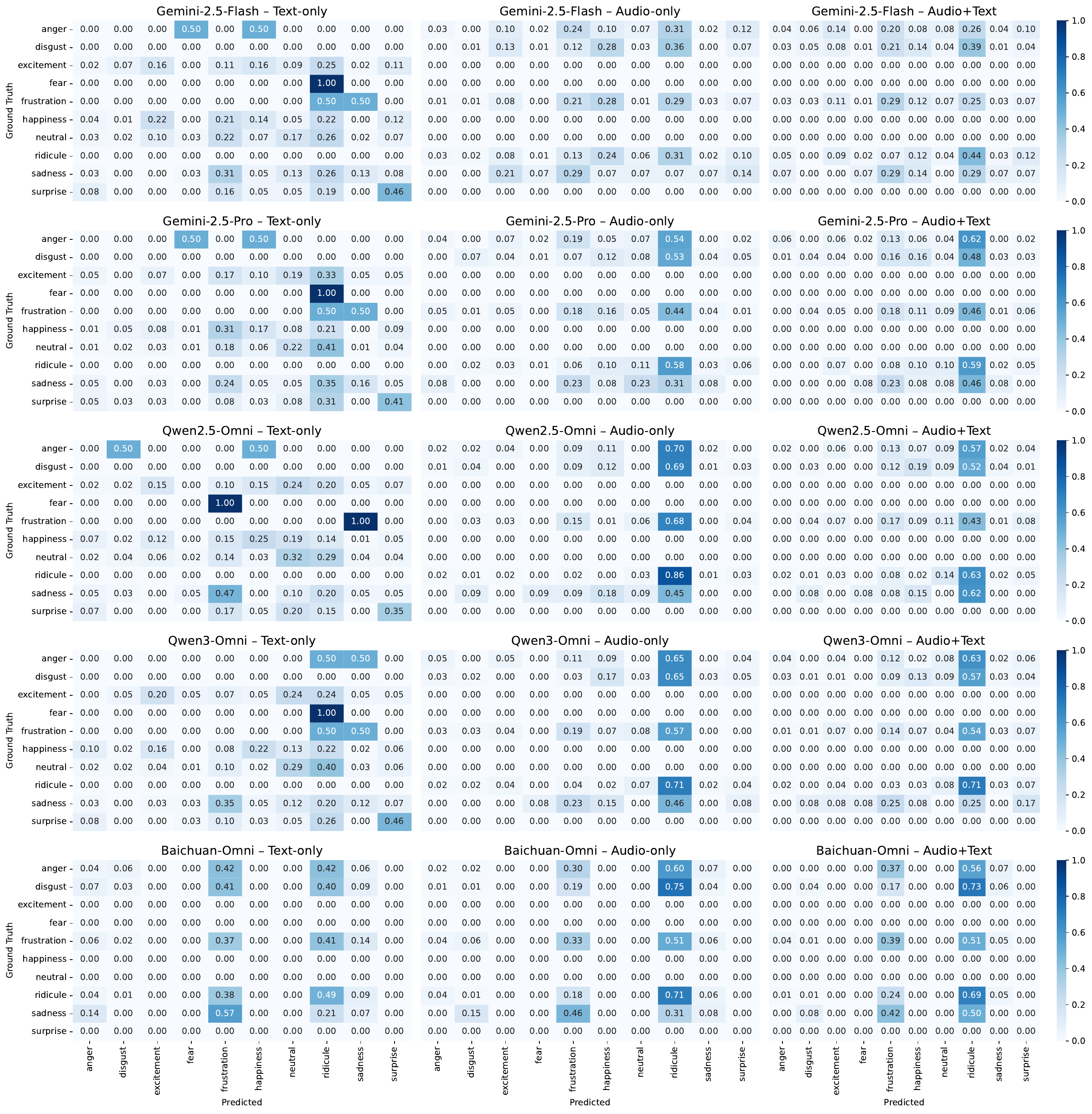}
  \caption{Confusion matrices showing all models' emotion recognition performance for Emotion-Mismatched condition in the LISTEN benchmark. Row-normalized matrices display prediction distributions for each tested with text-only, audio-only, and audio+text modalities.}
  \label{fig:all_ex3}
\end{figure*}

\begin{figure*}[t]
  \centering
  \includegraphics[ width=\linewidth]{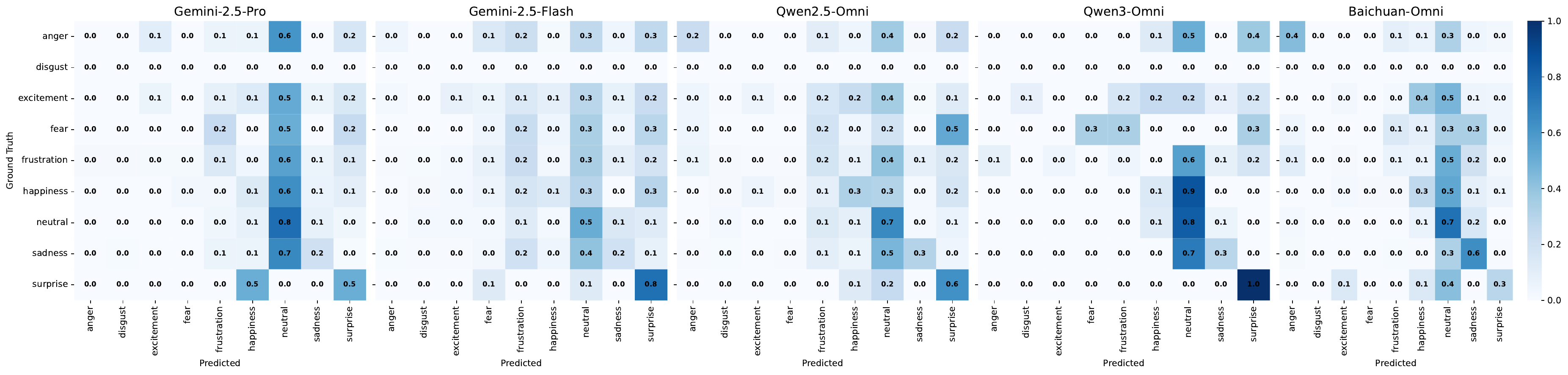}
  \caption{Confusion matrices showing all models' emotion recognition performance for Paralinguistic condition in the LISTEN benchmark. Row-normalized matrices display prediction distributions for each tested with audio-only modality.}
  \label{fig:all_exp4}
\end{figure*}

This presents the complete set of confusion matrices for all evaluated models across the three experimental conditions (Neutral-Text, Emotion-Matched, and Emotion-Mismatched) and three input modalities (Text, Audio, and Text+Audio). (see \autoref{fig:all_exp1}, \autoref{fig:all_exp2}, \autoref{fig:all_ex3}, \autoref{fig:all_exp4}) These figures provide a detailed view of each model’s prediction distribution across emotion categories, complementing the main results discussed in \autoref{sec:results}.


\newpage

\subsection{Condition Examples}  
\label{app:condition_examples}
We provides representative examples for each condition. (see \autoref{fig:1C}, \autoref{fig:2A}, \autoref{fig:2B}, \autoref{fig:2C}, \autoref{fig:3A}, \autoref{fig:3B}, \autoref{fig:3C}, and \autoref{fig:4} \autoref{app:1A} and \autoref{fig:1B})

\begin{figure*}[t]
\centering
\begin{tcolorbox}[colback=gray!10, colframe=gray!50, title=Neutral-Text Text-only, width=\textwidth, boxrule=0.5pt]
\begin{itemize}[leftmargin=*, label={}]
  \item \textbf{Modality:} Text-only
  \item \textbf{Sample ID:} \texttt{SAMPLE\_7c8b53fb}
  \item \textbf{Transcription:} \textit{"Kids are talking by the door."}
  \item \textbf{Prompt:} Read the transcription and classify the emotion.  
  Based on the content of this text, what emotion would the person likely be feeling?
  \begin{itemize}[leftmargin=2em, label={}]
    \item A. anger
    \item B. fear
    \item C. disgust
    \item D. neutral
    \item E. sadness
    \item F. surprise
    \item G. calm
    \item H. happiness
  \end{itemize}
  \item \textbf{Expected Response:} \textcolor{highlightred}{D (neutral)}
  \item \textbf{Model Prediction:} \textcolor{highlightred}{D (neutral)}
  \item \textbf{Ground Truth:} \textcolor{highlightred}{neutral}
  \item \textbf{Evaluation:}  \textcolor{choicegreen}{Correct}
\end{itemize}
\end{tcolorbox}
\caption{Example of a Neutral-Text text only entry in the LISTEN benchmark. The model correctly identifies the statement’s emotion (\textit{neutral}) when both lexical cues is available.}
\label{app:1A}
\end{figure*}

\begin{figure*}[t]
\centering
\begin{tcolorbox}[colback=gray!10, colframe=gray!50, title=Neutral-Text Audio-only, width=\textwidth, boxrule=0.5pt]
\begin{itemize}[leftmargin=*, label={}]
  \item \textbf{Modality:} Audio-only
  \item \textbf{Sample ID:} \texttt{SAMPLE\_7c8b53fb}
  \item \textbf{Audio:} RAVDESS\_train\_0333
  \item \textbf{Prompt:} Listen to the audio and classify the emotion.  
  What emotion does the speaker convey through their tone?
  \begin{itemize}[leftmargin=2em, label={}]
    \item A. surprise
    \item B. sadness
    \item C. fear
    \item D. anger
    \item E. calm
    \item F. happiness
    \item G. neutral
    \item H. disgust
  \end{itemize}
  \item \textbf{Expected Response:} \textcolor{highlightred}{D (anger)}
  \item \textbf{Model Prediction:} \textcolor{highlightred}{D (anger)}
  \item \textbf{Ground Truth:} \textcolor{highlightred}{anger}
  \item \textbf{Evaluation:} \textcolor{choicegreen}{Correct}
\end{itemize}
\end{tcolorbox}
\caption{Example of a Neutral-Text audio-only entry in the LISTEN benchmark. The model correctly infers the intended emotion (\textit{anger}) based on vocal prosody alone.}
\label{fig:1B}
\end{figure*}

\begin{figure*}[t]
\centering
\begin{tcolorbox}[colback=gray!10, colframe=gray!50, title=Neutral-Text Text+Audio, width=\textwidth, boxrule=0.5pt]
\begin{itemize}[leftmargin=*, label={}]
  \item \textbf{Modality:} Text+Audio
  \item \textbf{Sample ID:} \texttt{SAMPLE\_7c8b53fb}
  
  \item \textbf{Transcription:} \textit{"Kids are talking by the door."}
  \item \textbf{Audio:} RAVDESS\_train\_0333
  \item \textbf{Prompt:} Listen to the audio and read the transcription, then classify the emotion.  
  What emotion does the speaker convey through their tone?
  \begin{itemize}[leftmargin=2em, label={}]
    \item A. surprise
    \item B. sadness
    \item C. fear
    \item D. anger
    \item E. calm
    \item F. happiness
    \item G. neutral
    \item H. disgust
  \end{itemize}
  \item \textbf{Expected Response:} \textcolor{highlightred}{D (anger)}
  \item \textbf{Model Prediction:} \textcolor{highlightred}{D (anger)}
  \item \textbf{Ground Truth:} \textcolor{highlightred}{anger}
  \item \textbf{Evaluation:} \textcolor{choicegreen}{Correct}
\end{itemize}
\end{tcolorbox}
\caption{Example of a Neutral-Text text+audio entry in the LISTEN benchmark. The model correctly identifies the speaker’s emotion (\textit{anger}) when both lexical and prosodic cues are available.}
\label{fig:1C}
\end{figure*}

\begin{figure*}[t]
\centering
\begin{tcolorbox}[colback=gray!10, colframe=gray!50, title=Emotion-Matched Text-only, width=\textwidth, boxrule=0.5pt]
\begin{itemize}[leftmargin=*, label={}]
  \item \textbf{Modality:} Text-only
  \item \textbf{Sample ID:} \texttt{SAMPLE\_9b76ea7d}
  \item \textbf{Transcription:} \textit{"What the hell is this?"}
  \item \textbf{Prompt:} Read the transcription and classify the emotion.  
  From the semantic content alone, what emotion is being expressed?
  \begin{itemize}[leftmargin=2em, label={}]
    \item A. neutral
    \item B. sadness
    \item C. excitement
    \item D. frustration
    \item E. fear
    \item F. disgust
    \item G. happiness
    \item H. anger
    \item I. surprise
  \end{itemize}
  \item \textbf{Expected Response:} \textcolor{highlightred}{H (frustration)}
  \item \textbf{Model Prediction:} \textcolor{gray!70}{D (neutral)}
  \item \textbf{Ground Truth:} \textcolor{highlightred}{frustration}
  \item \textbf{Evaluation:} \textcolor{gray!70}{Incorrect}
\end{itemize}
\end{tcolorbox}
\caption{Example of an Emotion-Matched text-only entry from the LISTEN benchmark. The model misclassified an explicitly frustration utterance (\textit{"What the hell is this?"}) as \textit{neutral}, illustrating overgeneralization across semantically related negative emotions.}
\label{fig:2A}
\end{figure*}

\begin{figure*}[t]
\centering
\begin{tcolorbox}[colback=gray!10, colframe=gray!50, title=Emotion-Matched Audio-only, width=\textwidth, boxrule=0.5pt]
\begin{itemize}[leftmargin=*, label={}]
  \item \textbf{Modality:} Audio-only
  \item \textbf{Sample ID:} \texttt{SAMPLE\_dd0f6e9d}
  \item \textbf{Audio:} IEMOCAP\_Session5\_Ses05M\_script01\_1b\_F030
  
  \item \textbf{Prompt:} Listen to the audio and classify the emotion.  
  Based on the vocal expression, what emotion is the speaker feeling?
  \begin{itemize}[leftmargin=2em, label={}]
    \item A. frustration
    \item B. anger
    \item C. neutral
    \item D. excitement
    \item E. happiness
    \item F. surprise
    \item G. disgust
    \item H. fear
    \item I. sadness
  \end{itemize}
  \item \textbf{Expected Response:} \textcolor{highlightred}{A (frustration)}
  \item \textbf{Model Prediction:} \textcolor{highlightred}{A (frustration)}
  \item \textbf{Ground Truth:} \textcolor{highlightred}{frustration}
  \item \textbf{Evaluation:} \textcolor{choicegreen}{Correct}
\end{itemize}
\end{tcolorbox}
\caption{Example of an Emotion-Matched audio-only entry from the LISTEN benchmark. The model correctly interprets prosodic cues to identify the emotion as \textit{frustration}, showing sensitivity to vocal intensity and tone even without textual input.}
\label{fig:2B}
\end{figure*}

\begin{figure*}[t]
\centering
\begin{tcolorbox}[colback=gray!10, colframe=gray!50, title=Emotion-Matched Text+Audio, width=\textwidth, boxrule=0.5pt]
\begin{itemize}[leftmargin=*, label={}]
  \item \textbf{Modality:} Text+Audio
  \item \textbf{Sample ID:} \texttt{SAMPLE\_dd0f6e9d}
  \item \textbf{Audio:} IEMOCAP\_Session5\_Ses05M\_script01\_1b\_F030
  \item \textbf{Transcription:} \textit{"What the hell is this?"}
  \item \textbf{Prompt:} Listen to the audio and read the transcription, then classify the emotion.  
  Based on the vocal expression, what emotion is the speaker feeling?
  \begin{itemize}[leftmargin=2em, label={}]
    \item A. frustration
    \item B. anger
    \item C. neutral
    \item D. excitement
    \item E. happiness
    \item F. surprise
    \item G. disgust
    \item H. fear
    \item I. sadness
  \end{itemize}
  \item \textbf{Expected Response:} \textcolor{highlightred}{A (frustration)}
  \item \textbf{Model Prediction:} \textcolor{highlightred}{A (frustration)}
  \item \textbf{Ground Truth:} \textcolor{highlightred}{frustration}
  \item \textbf{Evaluation:} \textcolor{choicegreen}{Correct}
\end{itemize}
\end{tcolorbox}
\caption{Example of an Emotion-Matched text+audio entry from the LISTEN benchmark. The model correctly identifies \textit{frustration} when integrating both lexical and prosodic cues, demonstrating effective multimodal fusion under congruent emotional alignment.}
\label{fig:2C}
\end{figure*}

 \begin{figure*}[t]
\centering
\begin{tcolorbox}[colback=gray!10, colframe=gray!50, title=Emotion-Mismatched Text-only, width=\textwidth, boxrule=0.5pt]
\begin{itemize}[leftmargin=*, label={}]
  \item \textbf{Modality:} Text-only
  \item \textbf{Sample ID:} \texttt{SAMPLE\_955399e0}
  \item \textbf{Transcription:} \textit{"You're right, the party's fantastic. Please, tell me more. I haven't heard enough about it all week because hearing about that never gets old!"}
  \item \textbf{Prompt:} Read the transcription and classify the emotion.  
  What emotion is conveyed by the words in this statement?
  \begin{itemize}[leftmargin=2em, label={}]
    \item A. surprise
    \item B. excitement
    \item C. sadness
    \item D. disgust
    \item E. fear
    \item F. neutral
    \item G. anger
    \item H. happiness
    \item I. frustration
    \item J. ridicule
  \end{itemize}
  \item \textbf{Expected Response:} \textcolor{highlightred}{B (excitement)}
  \item \textbf{Model Prediction:} \textcolor{highlightred}{B (excitement)}
  \item \textbf{Ground Truth:} \textcolor{highlightred}{excitement (explicit emotion label)}
  \item \textbf{Evaluation:} \textcolor{choicegreen}{Correct}
\end{itemize}
\end{tcolorbox}
\caption{Example of an Emotion-Mismatched text-only entry from the LISTEN benchmark. Although the lexical content expresses \textit{excitement}, the corresponding audio (not shown) conveys \textit{ridicule}, highlighting the designed lexical–prosodic conflict characteristic of this condition.}
\label{fig:3A}
\end{figure*}

\begin{figure*}[t]
\centering
\begin{tcolorbox}[colback=gray!10, colframe=gray!50, title=Emotion-Mismatched Audio-only, width=\textwidth, boxrule=0.5pt]
\begin{itemize}[leftmargin=*, label={}]
  \item \textbf{Modality:} Audio-only
  \item \textbf{Sample ID:} \texttt{SAMPLE\_c52e71d0}
  \item \textbf{Audio:} MUStARD\_PRO\_1\_7575\_u\_3B
  \item \textbf{Prompt:} Listen to the audio and classify the emotion.  
  What emotion is communicated through the speaker's vocal prosody?
  \begin{itemize}[leftmargin=2em, label={}]
    \item A. disgust
    \item B. neutral
    \item C. ridicule
    \item D. frustration
    \item E. sadness
    \item F. anger
    \item G. excitement
    \item H. fear
    \item I. surprise
    \item J. happiness
  \end{itemize}
  \item \textbf{Expected Response:} \textcolor{highlightred}{F (anger)}
  \item \textbf{Model Prediction:} \textcolor{gray!70}{G (excitement)}
  \item \textbf{Ground Truth:} \textcolor{highlightred}{anger (implicit emotion label)}
  \item \textbf{Evaluation:} \textcolor{gray!70}{Incorrect}
\end{itemize}
\end{tcolorbox}
\caption{Example of an Emotion-Mismatched audio-only entry from the LISTEN benchmark. The lexical content is superficially positive (\textit{“You’re right, the party’s fantastic”}), but the prosody expresses irritation and \textit{anger}. The model incorrectly predicts \textit{excitement}, indicating difficulty in resolving sarcastic or contrastive vocal tone.}
\label{fig:3B}
\end{figure*}

\begin{figure*}[t]
\centering
\begin{tcolorbox}[colback=gray!10, colframe=gray!50, title=Emotion-Mismatched Text+Audio, width=\textwidth, boxrule=0.5pt]
\begin{itemize}[leftmargin=*, label={}]
  \item \textbf{Modality:} Text+Audio
  \item \textbf{Sample ID:} \texttt{SAMPLE\_c52e71d0}
  \item \textbf{Audio:} \texttt{MUStARD\_PRO\_1\_7575\_u\_3B}
  \item \textbf{Transcription:} \textit{"You're right, the party's fantastic. Please, tell me more. I haven't heard enough about it all week because hearing about that never gets old!"}
  \item \textbf{Prompt:} Listen to the audio and read the transcription, then classify the emotion.  
  What emotion is communicated through the speaker's vocal prosody?
  \begin{itemize}[leftmargin=2em, label={}]
    \item A. neutral
    \item B. disgust
    \item C. anger
    \item D. sadness
    \item E. excitement
    \item F. fear
    \item G. ridicule
    \item H. frustration
    \item I. happiness
    \item J. surprise
  \end{itemize}
  \item \textbf{Expected Response:} \textcolor{highlightred}{C (anger)}
  \item \textbf{Model Prediction:} \textcolor{gray!70}{E (excitement)}
  \item \textbf{Ground Truth:} \textcolor{highlightred}{anger (implicit emotion label)}
  \item \textbf{Evaluation:} \textcolor{gray!70}{Incorrect}
\end{itemize}
\end{tcolorbox}
\caption{Example of an Emotion-Mismatched text+audio entry from the LISTEN benchmark. Despite access to both modalities, the model incorrectly predicts \textit{excitement} instead of the intended \textit{anger}, suggesting overreliance on lexical positivity rather than prosodic dissonance—a hallmark challenge in sarcasm and irony understanding.}
\label{fig:3C}
\end{figure*}
\begin{figure*}[t]
\centering
\begin{tcolorbox}[colback=gray!10, colframe=gray!50, title=Paralinguistic Audio-only, width=\textwidth, boxrule=0.5pt]
\begin{itemize}[leftmargin=*, label={}]
  \item \textbf{Modality:} Audio-only
  \item \textbf{Sample ID:} \texttt{SAMPLE\_54df39ff}
  \item \textbf{Audio:} \texttt{IEMOCAP\_Session5\_Ses05F\_impro03\_F006}
  \item \textbf{Prompt:} Listen to the audio and classify the emotion.  
  What emotional tone is conveyed by the literal meaning of this statement?
  \begin{itemize}[leftmargin=2em, label={}]
    \item A. anger
    \item B. happiness
    \item C. fear
    \item D. sadness
    \item E. surprise
    \item F. frustration
    \item G. excitement
    \item H. disgust
    \item I. neutral
  \end{itemize}
  \item \textbf{Expected Response:} \textcolor{highlightred}{G (excitement)}
  \item \textbf{Model Prediction:} \textcolor{gray!70}{B (happiness)}
  \item \textbf{Ground Truth:} \textcolor{highlightred}{excitement}
  \item \textbf{Evaluation:} \textcolor{gray!70}{Incorrect}
\end{itemize}
\end{tcolorbox}
\caption{Example of a Paralinguistic audio-only entry from the LISTEN benchmark. The utterance contains only nonverbal laughter, labeled as \textit{excitement}. The model incorrectly classifies it as \textit{happiness}, revealing challenges in distinguishing subtle affective intent from nonverbal vocalizations.}
\label{fig:4}
\end{figure*}

\end{document}